\documentclass{article} 
\usepackage{iclr2025_conference,times}

\usepackage{amsmath,amsfonts,bm}

\def\eqref#1{equation~\ref{#1}}

\def\1{\bm{1}}

\DeclareMathAlphabet{\mathsfit}{\encodingdefault}{\sfdefault}{m}{sl}
\SetMathAlphabet{\mathsfit}{bold}{\encodingdefault}{\sfdefault}{bx}{n}

\usepackage{hyperref}
\usepackage{url}

\usepackage[utf8]{inputenc} 
\usepackage[T1]{fontenc}    
\usepackage{booktabs}       
\usepackage{amsfonts}       
\usepackage{nicefrac}       
\usepackage{microtype}      
\usepackage{xcolor}         

\newcommand{\typo}[1]{\textcolor{black}{#1}}
\newcommand{\symbolic}[1]{\textcolor{black}{#1}}

\usepackage{amsmath}
\usepackage{amssymb}
\usepackage{mathtools}
\usepackage{amsthm}

\usepackage{wrapfig}
\usepackage{xspace}
\usepackage{booktabs}
\usepackage{graphicx}
\usepackage{algorithm}
\usepackage{algorithmic}
\usepackage{float}
\usepackage{multirow}
\usepackage{pgfplots}
\usepackage{makecell}
\usepackage{cleveref}
\usepackage{listings}
\usepackage{titlesec}
\usepackage{enumitem}
\crefname{section}{§}{§§}
\Crefname{section}{§}{§§}

\newcommand{\model}{{RaDAgent}\xspace}
\newcommand{\elo}{{Elo-based Utility Learning}\xspace}

\title{Rational Decision-Making Agent with Learning Internal Utility Judgment}

\author{Yining Ye$^{1}\thanks{\ \ Indicates equal contribution.}$\hspace{0.5em}, Xin Cong$^{1*}\thanks{\ \  Corresponding author.}$\hspace{0.5em}, Shizuo Tian$^1$, Yujia Qin$^1$, Chong Liu$^1$, \\
\textbf{Yankai Lin$^2\dag$, Zhiyuan Liu$^{1}$, Maosong Sun$^{1}$} \\
$^1$Tsinghua University $^2$Renmin University of China\\
\\
\texttt{\{yeyn23,tsz21\}@mails.tsinghua.edu.cn} \\
\texttt{\{congxin1995,liuzy,sms\}@mail.tsinghua.edu.cn} \\
\texttt{yujiaqin16@gmail.com, tylch2@163.com, yankailin@ruc.edu.cn}}

\iclrfinalcopy 
\begin{document}

\maketitle

\begin{abstract}
   With remarkable advancements, large language models~(LLMs) have attracted significant efforts to develop LLM-based agents capable of executing intricate multi-step decision-making tasks.
   Existing approaches predominantly build upon the external performance measure to guide the decision-making process but the reliance on the external performance measure as prior is problematic in real-world scenarios, where such prior may be unavailable, flawed, or even erroneous.
   For genuine autonomous decision-making for LLM-based agents, it is imperative to develop rationality from their posterior experiences to judge the utility of each decision independently.
   %
   %
   In this work, we propose \textbf{\model} (\underline{Ra}tional \underline{D}ecision-Making \underline{Agent}), which fosters the development of its rationality through an iterative framework involving \textit{Experience Exploration} and \textit{Utility Learning}.
   Within this framework, \elo is devised to assign Elo scores to individual decision steps to judge their utilities via pairwise comparisons. 
   Consequently, these Elo scores guide the decision-making process to derive optimal outcomes.
   Experimental results on the Game of 24, WebShop, ToolBench and RestBench datasets demonstrate \model's superiority over baselines, achieving about $7.8\%$ improvement on average. 
   \typo{Besides, \model can also reduce costs (ChatGPT API calls), highlighting its effectiveness and efficiency. }
 
\end{abstract}
\section{Introduction}

The autonomous agent~\citep{Searle1969SpeechAA,Wooldridge1995IntelligentAT,Maes1994AgentsTR,Hendler1999IsTA}, as the long-standing pursuit of artificial intelligence~(AI), is expected to possess the ability to plan, make decisions, and take actions to accomplish complex tasks autonomously.
As large language models~(LLMs) have undergone rapid development, showcasing remarkable capabilities~\citep{openaichatgptblog,openai2023gpt4,touvron2023llama,touvron2023llama2,dubey2024llama}, many efforts have been devoted to developing LLM-based agent~\citep{richards2023auto,nakajima2023babyagi,agentgpt,wu2023autogen} to accomplish intricate multi-step decision-making tasks beyond traditional natural language processing~(NLP) applications~\citep{yao2022react,hao2023reasoning,yao2023tree,qin2023toolllm,chen2023agentverse,qian2024chatdev}.
Even with these strides, existing LLM-based agents require external performance measures as \textit{prior} to guide their decision-making process~\citep{yao2023tree,hao2023reasoning,sel2023algorithm,lv2024codeact}.
For instance, in Game of 24, which uses four numbers and basic arithmetic operations to obtain 24, \citet{yao2023tree} heuristically design a prompt to assess the possibility of each decision to reach 24 and then choose the higher one as decisions accordingly.
However, this manual-designed prompt may not provide accurate possibility, causing unreliable decision-making guidance.
\textbf{The reliance on the external performance measure restricts the adaptability in real-world scenarios as it may be unavailable, flawed, or even erroneous.}

When individuals make decisions, they not only rely on external measures but also draw upon their practical experience as \textit{posterior} to form a sense of internal rationality. 
Human rational decision-making evolves through a dynamic process of experiential learning, encompassing trial-and-error, reflection, and reinforcement~\cite{kahneman2013prospect}. 
By experimenting with different decisions and observing their outcomes, individuals learn to reinforce behaviors that yield favorable results.
This learning process involves reflective analysis, during which individuals critically evaluate their decision-making processes to identify biases and rectify mistakes. 
Through these iterations, individuals progressively enhance their decision-making by reinforcing behaviors that produce positive outcomes, thereby refining their judgment of the utility of each decision.
Finally, individuals derive internal utility judgment which serves as the basis for evaluating the effectiveness of decisions and identifying optimal solutions~\citep{Arrow1959RationalCF,Plott1973PATHIR,Kahneman2000ChoicesVA}.


To this end, we propose \textbf{\model} (\underline{Ra}tional \underline{D}ecision-Making \underline{Agent}) which learns internal utility judgment ability to achieve rationality for the agent.
In \model, the internal utility judgment is constructed based on an iterative framework:
(1) \textbf{Experience Exploration}:
Due to the complexity of real-world tasks, the solution space may be infinite, and it is challenging to find the optimal solution efficiently.
Hence, working in the Monte Carlo Tree Search~\citep{kocsis2006bandit,coulom2006efficient} fashion, \model would explore potential decisions with higher utilities to find better solutions as many as possible for the following utility learning.
(2) \textbf{Utility Learning}:
Given a series of solutions, \model should make comparisons between them to learn their utilities.
%
%
Due to the challenge of LLMs in directly providing accurate numerical utilities without prior, we design \elo which employs the Elo Rating system~\citep{elo1967proposed} to estimate the utilities through posterior comparison among explored solutions.
After multiple comparisons, the Elo scores would converge to an accurate value representing its actual utility in achieving the task.
Using the learned utilities as guidance, the exploration process focuses on discovering decisions with higher utilities. 
Consequently, the exploration of these enhanced decisions aids in further refining the utilities associated with each decision.
Through the iterative utility judgment learning, \model can assess the numerical utility of explored decisions and then can judge the highest utility to derive the best solution with the superior outcome.

To validate the effectiveness of our \model, we implement it based on ChatGPT~\citep{openaichatgptblog} and conduct extensive experiments on Game of 24~\citep{yao2023tree}, WebShop~\citep{yao2022webshop}, ToolBench~\citep{qin2023toolllm} and RestBench~\citep{song2023restgpt}, which contains intricate multi-step decision tasks involving diverse scenarios.
Experimental results demonstrate the superiority of our approach against several baselines by achieving about $7.8\%$ improvements on average to accomplish complex tasks.
Moreover, extensive analyses show that our approach not only delivers superior solutions but also achieves greater efficiency by reducing the number of ChatGPT API calls.

Our contributions are threefold:
(1) We propose \model, a rational decision-making agent that can construct its internal rationality to accomplish diverse real-world tasks, not relying on external performance measures.
(2) We devise \elo which can learn internal utility judgment for the agent by learning Elo scores for each decision, selecting the optimal solution with the highest utilities.
(3) Extensive experiments on the Game of 24, WebShop, ToolBench, and RestBench datasets demonstrate the effectiveness and efficiency of our proposed method against representative methods.
Our source code is released in \texttt{https://github.com/OpenBMB/RaD-Agent}.



\section{Related Work}

\textbf{Decision-Making Methods for LLM-based Agents}
Efficient and effective decision-making ability is fundamental for LLM-based agents to the attainment of specific objectives~\citep{yao2022react,yao2023tree,hao2023reasoning,besta2023graph,sel2023algorithm}. 
Although LLMs are pre-trained on a large-scale corpus which equips them with substantial common sense and knowledge to solve several problems, due to the complexity and diversity of realistic tasks, LLM-based agents still struggle to make multi-step decisions to solve realistic tasks.
Recently, as Chain-of-Thought~\citep{wei2023chainofthought} demonstrates its capability to decompose complex questions into sequential intermediate steps, several LLM-based decision-making methods have been proposed to enhance the decision-making ability of agents.
ReACT~\citep{yao2022react} develops a variant of CoT to leverage the reasoning ability of LLMs in decision-making scenarios.
Reflexion~\citep{shinn2023reflexion} further offers a remedial approach to make LLMs reflect their failure and summarize the reason in the decision process, and then correct their mistake in the second attempt. 
Based on these methods, some tree-based decision-making methods are proposed to adapt the decision-making ability of LLMs into specific tasks.
Tree-of-Thought~\citep{yao2023tree} proposes BFS and DFS decision-making algorithms in Game of 24, Creative Writing, and Mini Crosswords tasks.
RAP~\citep{hao2023reasoning} applies the Monte Carlo Tree search algorithm to find a good solution in Blocksworld, Math Reasoning, and Logical Reasoning tasks.
DFSDT~\citep{qin2023toolllm}, following a similar tree search algorithm, proposes an efficient version of DFS to make decisions.
However, the aforementioned methods need external performance measures to guide the decision-making process, which limits their scope of application.
In this paper, we propose \model which learns the internal utility judgment ability with the Elo rating system to achieve rationality for agents to provide optimal solutions.

\textbf{Tool Learning}
Recent investigations have cast illumination upon the burgeoning proficiencies exhibited by LLM-based agents in the mastery of instruments and the execution of decision-making processes within intricate contextual milieus~\citep{qin2023tool,vemprala2023chatgpt,nakano2021webgpt,qin2023webcpm,shen2023hugginggpt,wu2023visual,schick2023toolformer,hao2023toolkengpt,qian2023creator,song2023restgpt,qin2023toolllm,guo2024stabletoolbench}.
The incorporation of external tools into the operational framework of LLM-based agents confers upon them immediate access to contemporaneous factual knowledge~\citep{yang2023chatgpt}, imbues them with versatile multimodal capabilities~\citep{gupta2023visual}, and empowers them with specialized proficiencies tailored to vertical domains~\citep{jin2023genegpt}.
However, when confronted with real-world tasks that often require the utilization of multiple tools, LLM-based agents must engage in multi-step decision-making processes to select tools and determine their sequencing.
Consequently, the ability for decision-making in tool learning scenarios becomes imperative to effectively tackle practical tasks.

\section{Preliminaries}

\paragraph{Markov Decision Process}
\label{sec:task_formulation}
\symbolic{
We formulate the decision-making process within the agent as a finite-horizon Markov Decision Process~(MDP) denoted by $\mathcal{M} = \{ \mathcal{S}, \mathcal{A}, \mathcal{R}, \mathcal{T} \}$ with state space $\mathcal{S}$, action space $\mathcal{A}$, reward function $\mathcal{R}$, and transition function $\mathcal{T}$.
Given a human instruction $Q$, the agent acting as the policy model $\pi$ is tasked with generating a decision sequence to accomplish $Q$. 
The agent $\pi$ starts from the initial state $s_0$ and takes an action $a_i$ based on the current state and subsequently arrives at the next state $s_{i+1}$ decided by the transition function $\mathcal{T}$.
This process terminates until the agent accomplishes the task or exceeds the limitation of the number of actions, resulting in a decision sequence or trajectory $\tau = \{ s_0, a_1, s_1, \cdots, s_N \}$ where $N$ is the number of actions.
A reward of $1$ is assigned by the reward function $\mathcal{R}$ at the end if the agent successfully accomplishes the task, otherwise a reward of $0$ is assigned.
To make sequential decisions toward accomplishing $Q$ autonomously, we argue that the agent needs to identify the utility $v_{i}$ of each decision $a_{i}$ and select those decisions with a higher value that holds the promise of yielding the most promising outcomes (i.e., reward), ultimately leading to the derivation of the final decision sequence that fulfills the requirements of $Q$.
}


\paragraph{Elo Rating System}
The Elo rating system~\citep{elo1967proposed}, commonly used in competitive contexts offers a numerical estimation of the skill levels of players. 
It represents the skill levels of players by Elo scores and assesses the Elo scores through a series of one-to-one competitions.
%
%
It assumes that each player's performance follows a Gaussian distribution ($x \sim \mathcal{N}(\mu, \sigma)$) and each comparison of two players is actually comparing between two samples from their Gaussian distributions.
Through multiple comparisons, we can approximate their real skill levels by estimating their Elo scores.

Given two players $x$ and $y$, their Elo scores are denoted as $v_{x}$ and $v_{y}$, respectively.
The expected superiority of $x$ against $y$ is calculated as:
\begin{equation}
\label{eqn:Elo_expect}
    E_{x>y} = \frac{1}{1 + e^{-\frac{v_{x} - v_{y}}{r}}}
\end{equation}
where $r$ is the Elo coefficient. 

Next, we run a competition between them to find the actual winner.
We denote the result as $R_{x>y}$:
\begin{equation}
    R_{x>y} = \begin{cases}
    1, \text{if} \ x \ \text{win}, \\
    0, \text{if} \ y \ \text{win}, \\
    0.5, \text{otherwise}
    \end{cases}
\end{equation}
\typo{We then update their Elo scores accordingly:}
\begin{equation}
\label{eqn:Elo_update}
\begin{aligned}
    v_{x} = v_{x} + K * (R_{x>y} - E_{x>y}) \\
    v_{y} = v_{y} + K * (R_{y>x} - E_{y>x})
\end{aligned}
\end{equation}
where $K > 0$ is the update step size.
After multiple comparisons, the Elo score will progressively converge to their expected skill levels. 


\begin{algorithm}[t]
    \small
    \caption{RaDAgent}
    \label{alg:radagent}
    \begin{algorithmic}[1]
    \STATE \textbf{function} RaDAgent
    \STATE root $\leftarrow$ initialize an empty decision node
    \STATE \hspace{1em} \textbf{while} within computational budget \textbf{do}
    \STATE \hspace{2em} \texttt{\# Experience Exploration}
    \STATE \hspace{2em} node $\leftarrow$ root
    \STATE \hspace{2em} \textbf{while} node is not new-explored \textbf{do}
    \STATE \hspace{3em} node $\leftarrow$ sample node based on Equation~\ref{eqn:Elo_select}
    \STATE \hspace{2em} \textbf{end while}
    \STATE \hspace{2em} trace $\leftarrow$ generate new decision trace from node based on ReAct

    \STATE
    \STATE \hspace{2em} \texttt{\# Utility Learning}
    \STATE \hspace{2em} \textbf{while} within comparison limitation \textbf{do}
    \STATE \hspace{3em} candidata $\leftarrow$ sample an existing trace randomly
    \STATE \hspace{3em} result $\leftarrow$ compare new trace with candidate based on Equation~\ref{eqn:Elo_compare}
    \STATE \hspace{3em} update Elo scores based on result according to Equation~\ref{eqn:Elo_update}
    \STATE \hspace{3em} update exploration temperature $T$ based on Equation~\ref{eqn:Elo_anneal}
    \STATE \hspace{2em} \textbf{end while}
    \STATE \hspace{1em} \textbf{return} the best trace based on Equation~\ref{eqn:elo_final}
    \STATE \textbf{end function}



    \end{algorithmic}
\end{algorithm}

\section{Methodology}

Our \model aims to find the decision sequence with the highest utility to accomplish complex instructions autonomously.
It contains two principal phases to learn the internal utility judgment:
%
\begin{itemize}[topsep=1pt, partopsep=1pt, itemsep=-1pt, leftmargin=10pt]
    \item \textit{Experience Exploration}: The agent takes actions sequentially to form a decision sequence toward a feasible solution.
    \item \textit{Utility Learning}: The agent makes judgments among decision sequences to assess the utility (i.e., Elo scores) of existing decision steps.
\end{itemize}
These two phases work in an iterative fashion, reinforcing each other's outcomes (as shown in Algorithm~\ref{alg:radagent}).
In the experience exploration phase, the agent explores more potential decision sequences to find better solutions, which can encourage agents to learn the actual and accurate utility of each decision step.
In the utility learning phase, the agent re-calculates the Elo score of each decision step with the newly explored decision sequence to learn the utility of each decision, and the learned utilities serve as a dynamic guide, steering subsequent experience exploration toward more promising and superior solutions.
By iteratively cycling through these phases, the agent progressively evolves toward an optimal decision sequence with the highest utility to address instructions.

\subsection{Experience Exploration}
In \model, each experience exploration benefits from the previous exploration history based on \elo~(\cref{sec:decision_judgment}).
When exploring a new decision sequence, agents will select a decision step with a higher Elo score to explore further.
Specifically, in \model, each decision step is assigned an Elo score explicitly.
A decision step with higher Elo scores means that it is more likely to accomplish the instruction and thus Elo scores are used to guide the decision exploration process.
\symbolic{
Given an intermediate decision step $a$, its subsequent decision steps are denoted as $\{ a_1, a_2, \cdots, a_n \}$.
Given their learned Elo scores $\{ v_i \}_{i=1}^n$, the probability of choosing to explore can be modified as:
\begin{equation}
    P(a_i) = \frac{\exp({\frac{v_i}{T}})} {\sum_j \exp({\frac{v_j}{T}})}, \ a_i \in \{ a_1, a_2, \cdots, a_n \}
\end{equation}
where $T$ refers to the temperature. 
}
Note that only exploring the known decisions may cause the local optimal solution.
Therefore, we define a rejection decision step $\hat{a}$ with an initial Elo score $\hat{v}$ to represent that \textit{``The agent decides to explore a new decision''}.
\symbolic{
We add this rejection decision step into the subsequent decision steps as $\{ a_1, a_2, \cdots, a_n, \hat{a} \}$ when selecting:
\begin{equation}
\label{eqn:Elo_select}
    P(a_i) = \frac{\exp({\frac{v_i}{T}})} {\sum_j \exp({\frac{v_j}{T}})}, \ a_i \in \{ a_1, a_2, \cdots, a_n, \hat{a} \}
\end{equation}
}

The complete experience exploration process begins from the initial state $s_0$ and chooses the subsequent decision steps iteratively based on Equation~\ref{eqn:Elo_select} in a top-down manner.
When it chooses the rejection decision step $\hat{a}$, the agent will generate a new decision sequence starting from the intermediate step $a$.
In the iterative experience exploration process, those potential decision steps will be explored thoroughly, until finding the optimal solution.

\subsection{Utility Learning}
\label{sec:decision_judgment}
As external performance measures may be unavailable, flawed, or even erroneous, the agent should resort to their internal utility judgment ability to solve diverse tasks.
To this end, we design an \elo, equipping the agent with the Elo rating system to provide a numerical utility to each decision step to guide the decision-making process. 

The utility learning process (i.e., the Elo score estimation process) is conducted in a bottom-up manner.
It first adjusts the Elo scores of the final decision steps of each decision sequence via pairwise comparison and then updates the Elo scores of the intermediate decision steps gradually.
Once a new decision sequence is generated in the experience exploration phase, the agent will self-judge the Elo scores of existing decision steps via pairwise comparison.
\symbolic{
Given the newly generated decision sequence $\tau_{n}$, we first assign all decision steps of $\tau_{n}$ with an initial Elo score.
Then, we randomly select a decision sequence $\tau_{i}$ from existing decision sequences $\mathbb{T} = \{ \tau_1, \tau_2, \cdots, \tau_{n-1} \}$ and use agents to compare $\tau_{n}$ with $\tau_{i}$ to judge which one has the superior performance.
}
Since the LLM-based comparison is sensitive to the candidate order~\citep{qin2023large,chiang2023can,wang2023large}, we conduct comparisons twice with different orders.
\begin{equation}
    \label{eqn:Elo_compare}
    R_{t_{n}>t_{i}} = \begin{cases}
    1, \text{if} \ \tau_{n} \ \text{win twice}, \\
    0, \text{if} \ \tau_{i} \ \text{win twice}, \\
    0.5, \text{otherwise}
    \end{cases}
\end{equation}
Getting the comparison result, we update the Elo scores of the final decision steps of $\tau_{n}$ and $\tau_{i}$ based on Equation~\ref{eqn:Elo_update}.
Next, we calculate the Elo scores of intermediate decision steps based on their subsequent decision steps.
\symbolic{
Specifically, given an intermediate decision step $a_i$, we calculate its Elo scores as follows:
\begin{equation}
    v_{i} =  \sum_{a_j \in \text{Child}(a_i)} (\alpha_j * v_j),
\end{equation}
where $\text{Child}(a_i)$ refers to the set of the subsequent decision steps of $a_i$, $\alpha_j = \frac{\exp(v_j / T )}{\sum_{k} \exp(v_k / T)}$ is the normalized weight and $T$ is from Equation~\ref{eqn:Elo_select}.
}
By repeating the comparison via randomly sampling decision sequences, the Elo score of each decision step will converge to its expected value.

When guiding the experience exploration process, the Elo score of a decision step with a few number of Elo updates may not represent its real value accurately.
Such a decision step cannot be fully trusted for exhaustive exploration.
Hence, we adjust the temperature $T$ in Equation~\ref{eqn:Elo_select} based on the number of the Elo update.
\symbolic{
Let $M_{a}$ be the number of the Elo update of the decision step $a$.
The temperature of $a$ is annealed as follows:
\begin{equation}
    \label{eqn:Elo_anneal}
    T_{a} = T_0 * \frac{1}{1+\sqrt{\ln(M_{a}+1)}}
\end{equation}
where $T_0$ is the default temperature.
}
With the growth of the number of Elo updates, the approximated Elo score converges to its real value. 
At this time, we tend to explore the most possible decision.


After engaging in extensive experience exploration and utility learning, the agent learns the internal utility judgment to construct rationality that allows it to select the best-performed one as the final solution.
%
%
%
%
%
%
\symbolic{
Specifically, given all existing decision sequences $\mathbb{T} = \{ \tau_{1}, \tau_{2}, \cdots, \tau_{n} \}$, the one which final decision with the highest utility is selected as the final solution.
\begin{equation}
\label{eqn:elo_final}
t = \mathop{\arg\max}\limits_{\tau \in \mathbb{T}} \{ V(a_N)\}
\end{equation}
}
where $a_N$ refers to the final decision step.

\section{Experiments}

%

\subsection{Experimental Settings}
\label{sec:exp_settings}

\textbf{Datasets}
We conduct extensive experiments on Game of 24~\citep{yao2023tree}, WebShop~\citep{yao2022webshop}, and ToolBench~\citep{qin2023toolllm} datasets.
Game of 24 aims to use 4 numbers and four fundamental arithmetic operations ($+-*/$) to reach 24.
WebShop focuses on simulating the process of searching, browsing, and selecting items on an online shopping platform in order to obtain desired items.
ToolBench has thoughtfully constructed a diverse and intricate collection of human instructions of over $16$K APIs from $49$ categories.
We focused on the intra-category multi-tool instruction scenario which accurately reflects the complexities involved in real-world tasks, necessitating the use of various tools and requiring multi-step decision-making processes.
\typo{
We use $100$, $500$, and $500$ instances for Game of 24, WebShop, and ToolBench to evaluate the decision-making ability respectively.
}
Details of each task (including task description, action space, etc) can refer to \Cref{sec:task_description}.

\textbf{Baselines}
We compare \model with the following decision-making methods:
(1) \textbf{CoT}~\citep{wei2023chainofthought,yao2022react} decomposes reasoning into explicit intermediate steps and we adapt ReAct~\citep{yao2022react} to make sequential decisions.
\typo{
(2) \textbf{CoT@3} extends CoT by running the decision-making process three times independently for an instruction and finally generates a total of three decision sequences.
}
(3) \textbf{Reflexion}~\citep{shinn2023reflexion} builds upon CoT@3 and allows LLMs to engage in self-reflection on their previous decision sequences. 
(4) \textbf{ToT-BFS}~\citep{yao2023tree} constructs a decision tree in a top-down manner to search for a feasible solution. 
(5) \textbf{ToT-DFS}~\citep{yao2023tree} constructs a decision tree by going as deep as possible along each branch and exploring the most recently visited states. 
(6) \textbf{DFSDT}~\citep{qin2023toolllm} is an improved version of DFS, which allows agents to dynamically assess different decision states and choose to either proceed along a promising path or abandon an existing state and expand another one. 

\textbf{Evaluation Metrics}
\label{sec:metrics}
To ensure a rigorous and accurate evaluation of the performance of our proposed decision-making approach, we adopt three evaluation metrics for each dataset respectively:
(1) \textbf{Success Rate}~\citep{yao2023tree} measures the proportion of valid equations generated by the agent's arithmetic operations that yield a result of 24, using the given input numbers.
(2) \textbf{Reward}~\citep{yao2022webshop} evaluates the similarity (a value between $0$ and $1$) between the attributes of the items chosen by the agent and the attributes of the items actually purchased by human. 
(3) \textbf{Pass Rate}~\citep{qin2023toolllm} assesses the ability of agents to successfully accomplish complex real-world tasks by using tools sequentially. 
It calculates the proportion of instructions that an agent completes.

\textbf{Implementation Details}
\label{sec:implementation}

We use OpenAI ChatGPT \texttt{gpt-3.5-turbo-0613-16k} to implement our approach (our designed prompt can refer to \Cref{sec:prompt}). 
Our approach involves conducting a decision-exploration process $20$ times and finally selecting the decision sequence with the highest Elo score as the final decision.
For \elo, the initial Elo score of the decision step is set as $0.0$ and the Elo coefficient $r$ is set as $173.72$ according to the vanilla Elo rating system~\citep{elo1967proposed}. 
The Elo score of $\hat{d}$ in Equation~\ref{eqn:Elo_select} is set as $0.0$.
$K$ in Equation~\ref{eqn:Elo_update} is set as $50$.
%
%
To manage the computational cost of ChatGPT API calls, we set a maximum limit of $12$ steps for each decision sequence for a decision-searching process. 
Detailed analyses of each hyperparameter are discussed in \Cref{sec:hyperparameter}.

\begin{table*}[!t]
    \centering
    \begin{minipage}{0.5\linewidth}
        \small
        \centering
        \caption{Main experimental results on Game of 24, WebShop, and ToolBench dataset. Bold marks the best performance.}
        \vskip 0.1in
        \begin{tabular}{l|r|r|r}
            \toprule
            Model & Game of 24 & WebShop & ToolBench \\
            \midrule
             CoT & 6.00 & 56.23 & 16.60\\
             CoT@3 & 7.00 & 56.45 & 31.20\\
             Reflexion & 7.00 & 57.21 & 26.60\\
             ToT-BFS & 11.00 & 50.20 & 38.00\\
             ToT-DFS & 14.00 & 55.60 & 45.58\\
             DFSDT & 29.00 & 57.25 &  50.20\\
             \midrule
             \model & \textbf{43.00} & \textbf{59.36} & \textbf{61.92} \\
             \bottomrule
        \end{tabular}
        \label{tab:main}
    \end{minipage}%
    \hfill
    \begin{minipage}{0.46\linewidth}
        \small
        \centering
        \caption{Results on the real-world RestBench dataset. Baselines are reported from \citet{song2023restgpt}. Bold marks the best performance.} %
        \vskip 0.1in
        \begin{tabular}{l|r|r}
            \toprule
            Model & TMDB & Spotify\\
            \midrule
             Offline & 33.0 & 36.4\\ 
             DEPS & 43.0 & 43.8 \\
             ReAct & 57.0 & 49.1 \\ 
             Reflexion & 59.0 & 61.4 \\ 
             RestGPT & 79.0 & 74.5 \\
             RestGPT(ChatGPT) & 65.0 & 72.3 \\
             \midrule
             \model & \textbf{84.0} & \textbf{80.7} \\
             \bottomrule
        \end{tabular}
        \label{tab:restbench}
    \end{minipage}%
    
\end{table*}

\subsection{Overall Results}

To validate the effectiveness of our proposed \model approach, we first study whether our approach can accomplish more complex tasks.
The results are shown in Table~\ref{tab:main} and we can observe that:
(1) Across all datasets, \model consistently outperforms all baselines, indicating that incorporating the utility judgment internally empowers agents to accomplish a broader range of tasks effectively.
(2) In Game of 24\footnote{Note that the performance of Game of 24 is reproduced based on the \texttt{gpt-3.5-turbo-0613-16k} which is inconsistent with the reported results in the official paper.} and ToolBench domains, \model exhibits the capability to assign lower elo scores to decision steps that lead to failure. 
Consequently, these Elo scores serve as guidance for agents to avoid such decisions and achieve success.
(3) For Webshop, while our method still outperforms all baselines, it only achieves only marginal gains. 
This is attributed to the fact that Webshop provides only one ``golden answer'' for each instruction, while several other items actually meet the requirements. 
Consequently, these alternative items receive lower rewards as they deviate from the golden answer, resulting in an underestimation of the performance.

\section{Analysis}

\subsection{Generalization To Real-World Environment}
To verify that our method is robust and applicable to real-world environments, we expand our evaluation to the RestBench dataset~\citep{song2023restgpt}, which features restful APIs from two prominent real-world applications: TMDB and Spotify. 
All APIs in RestBench will be authentically called and get the \textbf{real-time}, \textbf{dynamic}, and \textbf{unpredictable} responses from the TMDB and Spotify platform, revealing a more challenging scenario for decision-making.
This dataset includes human-annotated real-world tasks with ground truth decision sequences. 
We compare our method with Offline~\citep{qin2023tool}, DEPS~\citep{wang2023describe}, ReAct~\citep{yao2022react}, Reflection~\citep{shinn2023reflexion}, RestGPT~\citep{song2023restgpt} and its ChatGPT version.
Following RestGPT~\citep{song2023restgpt}, we report the \textbf{Correct Path Rate} which calculates the proportion of the correct decision sequences.
Experimental results are shown in \Cref{tab:restbench}, which have demonstrated that our approach outperforms the baselines and achieves the best Correct Path Rate ($84.0\%$ and $80.7\%$ for TMDB and Spotify respectively). 
This result underscores the effectiveness of our method in making decisions in real-world environments with real-time, dynamic, and unpredictable features.

\subsection{Efficiency Analysis}
\label{sec:efficiency}

\begin{figure*}[!t]
	\centering
	\begin{minipage}[t]{0.45\linewidth}
        \centering
		\includegraphics[width=1.0\linewidth]{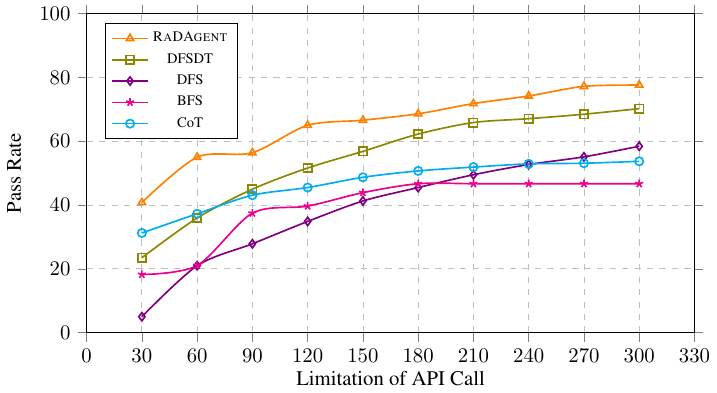}
		\vspace{-15pt}
        \caption{Efficiency results.}
		\label{fig:efficiency}
	\end{minipage}
	\hfill
	\begin{minipage}[t]{0.45\linewidth}
        \centering
		\includegraphics[width=1.0\linewidth]{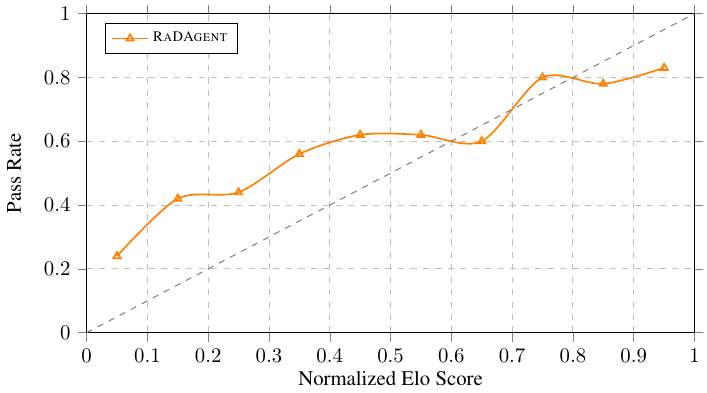}
        \vspace{-15pt}
		\caption{Performance with varied Elo scores.}
		\label{fig:elo}
	\end{minipage}
\end{figure*}

We further conducted the analyses to evaluate the efficiency of our proposed \model.
As making a decision step will involve a ChatGPT API call, the inefficient decision-making method would involve more API calls to accomplish the same instruction, causing costly expenses.
\typo{
We thus conducted experiments with varying ChatGPT API call limitations, ranging from $30$ to $300$, and measured the Pass Rate in ToolBench of each method under these varied limitations.
}
The experimental results are demonstrated in Figure~\ref{fig:efficiency}.
\typo{
These results showcase that the ToT-BFS, ToT-DFS, and DFSDT heavily rely on a large number of ChatGPT API calls to achieve a high Pass Rate.
}
Once limiting the number of API calls, their performance even cannot surpass CoT.
In contrast, our approach achieves the highest Pass Rate under all limitation settings, especially in low-resource settings.
We attribute it to the fact that our method can utilize Elo scores to dynamically select the promising decision steps, avoiding those unpromising ones.
Thus, our method illustrates superior efficiency against baselines and the practical advantages of our approach in real-world scenarios.
\typo{
To validate our generalized efficiency advantage, We have further conducted detailed efficiency analyses on Game of 24 and the complete experimental results are presented in \Cref{sec:efficiency_on_24}.
}

\subsection{Solution Ranking}

\begin{wraptable}{r}{0.4\linewidth}
    \small
    \centering
    \vspace{-21pt}
    \caption{Solution ranking experimental results on ToolBench dataset. Bold marks the best (the smaller, the better).}
    \vskip 0.1in
    \begin{tabular}{l|r}
        \toprule
        Model & Pref. Rank \\
        \midrule
         CoT@3 & 3.45 \\ 
         Reflexion & 3.48 \\
         ToT-BFS & 3.25 \\ 
         DFSDT & 2.91 \\ 
         \midrule
         \model & \\
         \quad w/ \textit{RandSelect} & 3.24 \\
         \quad w/ \textit{EloSelect} & \textbf{2.19} \\ 
         \bottomrule
    \end{tabular}
    \label{tab:rank}
\end{wraptable}
In addition to validating the effectiveness of our approach to reach feasible solutions, we seek to investigate whether \model can further provide solutions with higher quality. 
We adopt \textit{Win Rate} in ToolEval from ToolBench to compare the decision sequences produced by different methods for a given instruction.
Based \textit{Win Rate}, we utilize PRP~\citep{qin2023large} to rank decision sequences of all methods to report their rank to measure the superiority of each decision sequence. 
We further develop a variant of our model named \textit{RandSelect} which selects the final decision sequence randomly while \textit{EloSelect} selects based on the highest Elo score.
We then select representative baselines (CoT@3, Reflexion, ToT-BFS, ToT-DFS, DFSDT) and conduct a comprehensive comparison of the decision sequences produced by each method.
The experimental results are summarized in Table~\ref{tab:rank}, and it reveals that \model consistently achieves the top rank ($2.19$ on average) among all comparable baselines.
Especially, \textit{EloSelect} obviously outperforms \textit{RandSelect}, confirming the capability of our \elo to assess the utility of each decision sequence to select superior solutions, resulting in high-quality decision-making.

\subsection{Calibration Property of Elo-based Utility}

To verify the effectiveness of our \elo in providing reliable utility assessments, we conducted a comprehensive analysis using the ToolBench dataset. 
As the Elo score serves as a metric to represent the utility of each decision, we seek to determine whether the Elo score is a reliable indicator of decision utility.
To this end, we partitioned the ToolBench dataset into several subsets based on the Elo scores assigned to the decision sequences generated by \model. 
We first collected the Elo scores for all decision sequences predicted by \model in ToolBench data and then normalized them to scale within the range of 0 to 1. 
Next, we sorted the normalized Elo scores and divided them into 10 intervals, getting 10 subsets of ToolBench data accordingly and calculated the Pass Rate for each subset.
\Cref{fig:elo} illustrates the experimental results. 
A discernible trend is observed: the Pass Rate consistently increases with higher Elo scores.
A higher Elo score indicates that the decision sequence is more likely to represent an accomplished solution to the instruction, whereas a lower Elo score suggests that the instruction may be more challenging, and the corresponding decision sequence may not effectively solve the instruction.
This clear positive correlation between the Elo score and the Pass Rate demonstrates the efficacy of the \elo in providing reliable assessments of decision utility.

\begin{table*}[!t]
    \centering
    \begin{minipage}{0.5\linewidth}
        \small
        \centering
        \caption{Success rate of different decision measures on Game of 24.}
        \vskip 0.1in
        \begin{tabular}{lr}
        \toprule
        Method & Success Rate \\
        \midrule
        DFSDT & 29.0 \\
        Handcrafted Measure & 26.0 \\
        LLM Measure (\model) & 43.0 \\
        \bottomrule
        \end{tabular}
        \label{tab:manual}
    \end{minipage}%
    \hfill
    \begin{minipage}{0.40\linewidth}
        \small
        \centering
        \caption{Comparison with MCTS variants on Game of 24.}
        \vskip 0.1in
        \begin{tabular}{lr}
        \toprule
        Method & Success Rate \\
        \midrule
        MCTS@40 & 22.0 \\
        MCTS@100 & 40.0 \\
        \model & 43.0 \\
        \bottomrule
        \end{tabular}
        \label{tab:mcts}
    \end{minipage}%
    
\end{table*}

\subsection{Reliability of LLM Evaluation}

As we utilize LLM itself to provide decision sequence comparisons instead of handcrafted external measures, we have conducted additional experiments to validate if \model can provide more reliable measures than handcrafted external measures. 
We heuristically design a decision measure strategy, instead of LLM, for comparing decision sequences for Game of 24 to guide the decision-making process. 
The strategy is to compare two decisions to decide which result is close to 24 (i.e., the difference between the final calculation result of four numbers and 24). 
The closer one wins. 
If their difference are the same, they are tied. 
We replace the LLM comparison with the manual-designed strategy to conduct the experiments.
The results are shown in \Cref{tab:manual}.
The performance of this handcrafted strategy (achieving only $26.0\%$ Success Rate) is notably inferior to \model even DFSDT. 
This handcrafted strategy does not reflect the real performance since the result of an arithmetical expression is close to 24 numerically does not mean it is close to 24 operationally.
The inaccurate external measure will mislead the decision-making procedure to inferior performance.
%
%

\subsection{Reliability of Elo Rating}

As we utilize the Elo rating system to learn the internal utilities, it is essential to assess the reliability of the Elo Rating system as a utility evaluation tool.
To this end, we have implemented the Monte Carlo Tree Search~(MCTS) baseline with a value evaluation algorithm proposed by ToT~\citep{yao2023tree} for Game of 24. 
Different from our method utilizing the Elo rating system, this baseline calculates the score at each decision step based on the value evaluation proposed by ToT and uses the standard \textit{Upper Confidence Bounds for Trees} algorithm~(UCT)~\citet{kocsis2006bandit} to guide the decision-making process. 
We introduced two variants: MCTS@40 and MCTS@100, by conducting 40 and 100 simulations at each decision step, respectively.
The outcomes of these experiments are presented in \Cref{tab:mcts}. 
It is evident that none of these MCTS variants could outperform \model. 
Despite MCTS@100 displaying a performance somewhat closer to \model, it necessitated 100 simulations for each decision step, leading to a significant number of API calls. 
Conversely, \model required only 20 exploratory steps to achieve superior performance, which can be attributed to the Elo scores' ability to provide precise directions for exploration guidance, which can thereby enhance both the effectiveness and efficiency of the decision-making process.

\subsection{Impact of Elo Update Step}

To validate the impact of the Elo update step $K$ in the Elo rating system, we conducted a series of experiments with different values of $K= \{ 10, 50, 100 \}$ on the Game of 24 scenario. 
The experimental results are listed in \Cref{tab:perf_of_diff_K}.
Through these experiments, we observed that $K=50$ yielded the most optimal performance for our \model. 
It is important to note that $K$ in the Elo update algorithm functions analogously to the learning rate in Stochastic Gradient Descent optimization algorithms~\citep{sra2011optimization}. 
The choice of $K$ significantly influences the rate during Elo scores converge to their accurate values. 
A larger $K$ may lead to instability in the Elo scores, as it causes larger adjustments, thereby potentially overshooting the optimal value. 
Conversely, a smaller $K$ can result in slower convergence, necessitating more comparisons to reach an accurate assessment. 
Our experimental results have shown that setting $K$ as $50$ can derive the best performance.

\subsection{Impact of Utility Learning Prompt}

We have conducted additional experiments to validate the impact of the prompt design in the Elo-based utility learning algorithm on the Game of 24 scenario. 
%
%
In this setting, we employed a more straightforward utility learning prompt to compare two decision sequences. 
This prompt is detailed in \Cref{sec:simple_prompt} and the experimental results are shown in Table~\ref{tab:perf_of_diff_method}.
From the results, we can observe that there was an obvious decrease in performance with the simpler prompt compared with our original \model but it still outperforms the best baseline DFSDT. 
\typo{
These results highlight a couple of key points:
}
(1) Impact of Prompt Design: The experiment demonstrated that the design of the utility learning prompt does indeed impact the performance of the system. 
A more complex or carefully crafted prompt contributes to better utility assessment, leading to more effective decision-making.
(2) Robustness of Utility Learning: Despite the reduced performance with a simpler prompt, the fact that \model continued to outperform the baseline indicates the inherent robustness of our utility learning approach. 
It suggests that while the prompt design is significant, the core mechanics of our Elo-based utility learning algorithm are strong enough to maintain a competitive edge even under suboptimal conditions.
These findings unveil the need for further research into the optimal design of utility learning prompts. 
We plan to explore a broader range of prompt complexities and styles to fully understand their impact on the efficacy of the utility learning process in the future.

\typo{
Additionally, to further verify the reliability of our \elo algorithm, we conduct \textbf{detailed hyperparameter analysis experiments in \Cref{sec:hyperparameter}} including the initial Elo score, the number of decision comparisons, initial decision solution, etc.
}

\begin{table*}[!t]
    \centering
    \begin{minipage}{0.45\linewidth}
        \small
        \centering
        \caption{Success rate of different $K$ in Elo score update on Game of 24.}
        \vskip 0.1in
        \begin{tabular}{lc}
        \toprule
        Value of $K$ & Success Rate \\
        \midrule
        $K=10$ & 34.0 \\
        $K=50$ & 48.0 \\
        $K=100$ & 43.0 \\
        \bottomrule
        \end{tabular}
        \label{tab:perf_of_diff_K}
    \end{minipage}%
    \hfill
    \begin{minipage}{0.45\linewidth}
        \small
        \centering
        \caption{Success rate of different prompts of the utility learning on Game of 24.}
        \vskip 0.1in
        \begin{tabular}{lc}
        \toprule
        Method & Success Rate \\
        \midrule
        DFSDT & 29.0 \\
        SimplePrompt & 36.0 \\
        ElaboratePrompt(\model) & 43.0 \\
        \bottomrule
        \end{tabular}
        \label{tab:perf_of_diff_method}
    \end{minipage}
\end{table*}

\subsection{Impact of Large Language Models}

\begin{wraptable}{r}{0.4\linewidth}
    \small
    \centering
    \vspace{-21pt}
    \caption{Success rate of \texttt{GPT-3.5} and \texttt{GPT-4} on Game of 24. ToT is reported from its original paper~\citep{yao2023tree}.}
    \vskip 0.1in
    \begin{tabular}{llr}
    \toprule
    LLM & Method & Success Rate \\
    \midrule
    \multirow{2}*{GPT-3.5} & ToT & 19.0 \\
    & \model & 43.0 \\
    \midrule
    \multirow{2}*{GPT-4} & ToT & 74.0 \\
    & \model & 82.0 \\
    \bottomrule
    \end{tabular}
    \label{tab:perf_of_diff_llm}
\end{wraptable}
To validate the effectiveness of different LLMs, we have conducted additional experiments integrating \texttt{GPT-4} into our \model instead of \texttt{GPT-3.5} on the Game of 24 scenario. 
The experimental results are shown in Table~\ref{tab:perf_of_diff_llm}.
We found that \model, leveraging the enhanced capabilities of \texttt{GPT-4}, demonstrates its superiority over its \texttt{GPT-3.5} version. 
This finding underscores the scalability and adaptability of \model. 
These results suggest the following:
(1) As LLMs continue to evolve, our model can capitalize on these advancements to further enhance decision-making efficiency and accuracy.
%
%
(2) Compared with the \texttt{GPT-4} version of ToT~\cite{yao2023tree}, integrating our decision-making approach continues to yield performance improvements, demonstrating the necessity of robust decision-making strategies even when employing advanced LLMs.
%
%
In summary, LLMs and decision-making approaches are complementary, mutually enhancing each other to achieve superior performance outcomes.

\section{Conclusion}

In this work, we have introduced \model to learn the internal utility judgment ability for agents to achieve rationality across a diverse range of real-world tasks. 
We design an iterative framework involving Experience Exploration and Utility Learning to enhance agents to learn numeric utility for each decision step and guide the decision-making process. 
Extensive experiments on the Game of 24, WebShop, ToolBench, and RestBench datasets have confirmed the effectiveness of \model, outperforming baseline methods by achieving notable improvements and producing higher-quality solutions. 
Moreover, the reduction in LLM API calls showcases the efficiency gains of our approach. 
By empowering agents with rationality, our work paves the way for their broader utilization in real-world scenarios, alleviating the reliance on external performance measures. 

\section*{Acknowledgements}
This work is supported by the Postdoctoral Fellowship Program of CPSF (Grant No. GZB20230343) and China Postdoctoral Science Foundation (Grant No. 2023M741945).

\bibliography{ref}

\begin{thebibliography}{52}
\providecommand{\natexlab}[1]{#1}
\providecommand{\url}[1]{\texttt{#1}}
\expandafter\ifx\csname urlstyle\endcsname\relax
  \providecommand{\doi}[1]{doi: #1}\else
  \providecommand{\doi}{doi: \begingroup \urlstyle{rm}\Url}\fi

\bibitem[AgentGPT(2023)]{agentgpt}
AgentGPT.
\newblock Agentgpt.
\newblock \emph{Python. https://github.com/reworkd/AgentGPT}, 2023.

\bibitem[Arrow(1959)]{Arrow1959RationalCF}
K.~Arrow.
\newblock Rational choice functions and orderings1.
\newblock \emph{Economica}, 26:\penalty0 121, 1959.

\bibitem[Besta et~al.(2023)Besta, Blach, Kubicek, Gerstenberger, Gianinazzi, Gajda, Lehmann, Podstawski, Niewiadomski, Nyczyk, et~al.]{besta2023graph}
Maciej Besta, Nils Blach, Ales Kubicek, Robert Gerstenberger, Lukas Gianinazzi, Joanna Gajda, Tomasz Lehmann, Michal Podstawski, Hubert Niewiadomski, Piotr Nyczyk, et~al.
\newblock Graph of thoughts: Solving elaborate problems with large language models.
\newblock \emph{arXiv preprint arXiv:2308.09687}, 2023.

\bibitem[Chen et~al.(2023)Chen, Su, Zuo, Yang, Yuan, Chan, Yu, Lu, Hung, Qian, et~al.]{chen2023agentverse}
Weize Chen, Yusheng Su, Jingwei Zuo, Cheng Yang, Chenfei Yuan, Chi-Min Chan, Heyang Yu, Yaxi Lu, Yi-Hsin Hung, Chen Qian, et~al.
\newblock Agentverse: Facilitating multi-agent collaboration and exploring emergent behaviors.
\newblock In \emph{The Twelfth International Conference on Learning Representations}, 2023.

\bibitem[Chiang \& Lee(2023)Chiang and Lee]{chiang2023can}
Cheng-Han Chiang and Hung-yi Lee.
\newblock Can large language models be an alternative to human evaluations?
\newblock \emph{arXiv preprint arXiv:2305.01937}, 2023.

\bibitem[Coulom(2006)]{coulom2006efficient}
R{\'e}mi Coulom.
\newblock Efficient selectivity and backup operators in monte-carlo tree search.
\newblock In \emph{International conference on computers and games}, pp.\  72--83. Springer, 2006.

\bibitem[Dubey et~al.(2024)Dubey, Jauhri, Pandey, Kadian, Al-Dahle, Letman, Mathur, Schelten, Yang, Fan, et~al.]{dubey2024llama}
Abhimanyu Dubey, Abhinav Jauhri, Abhinav Pandey, Abhishek Kadian, Ahmad Al-Dahle, Aiesha Letman, Akhil Mathur, Alan Schelten, Amy Yang, Angela Fan, et~al.
\newblock The llama 3 herd of models.
\newblock \emph{arXiv preprint arXiv:2407.21783}, 2024.

\bibitem[Elo(1967)]{elo1967proposed}
AE~Elo.
\newblock The proposed uscf rating system, its development, theory, and applications. chess life xxii (8): 242--247, 1967.

\bibitem[Elo \& Sloan(1978)Elo and Sloan]{elo1978rating}
Arpad~E Elo and Sam Sloan.
\newblock The rating of chessplayers: Past and present.
\newblock \emph{(No Title)}, 1978.

\bibitem[Guo et~al.(2024)Guo, Cheng, Wang, Liang, Qin, Li, Liu, Sun, and Liu]{guo2024stabletoolbench}
Zhicheng Guo, Sijie Cheng, Hao Wang, Shihao Liang, Yujia Qin, Peng Li, Zhiyuan Liu, Maosong Sun, and Yang Liu.
\newblock Stabletoolbench: Towards stable large-scale benchmarking on tool learning of large language models.
\newblock \emph{arXiv preprint arXiv:2403.07714}, 2024.

\bibitem[Gupta \& Kembhavi(2023)Gupta and Kembhavi]{gupta2023visual}
Tanmay Gupta and Aniruddha Kembhavi.
\newblock Visual programming: Compositional visual reasoning without training.
\newblock In \emph{Proceedings of the IEEE/CVF Conference on Computer Vision and Pattern Recognition}, pp.\  14953--14962, 2023.

\bibitem[Hao et~al.(2023{\natexlab{a}})Hao, Gu, Ma, Hong, Wang, Wang, and Hu]{hao2023reasoning}
Shibo Hao, Yi~Gu, Haodi Ma, Joshua~Jiahua Hong, Zhen Wang, Daisy~Zhe Wang, and Zhiting Hu.
\newblock Reasoning with language model is planning with world model.
\newblock \emph{arXiv preprint arXiv:2305.14992}, 2023{\natexlab{a}}.

\bibitem[Hao et~al.(2023{\natexlab{b}})Hao, Liu, Wang, and Hu]{hao2023toolkengpt}
Shibo Hao, Tianyang Liu, Zhen Wang, and Zhiting Hu.
\newblock Toolkengpt: Augmenting frozen language models with massive tools via tool embeddings.
\newblock \emph{arXiv preprint arXiv:2305.11554}, 2023{\natexlab{b}}.

\bibitem[Hendler(1999)]{Hendler1999IsTA}
J.~Hendler.
\newblock Is there an intelligent agent in your future? nature.
\newblock 1999.

\bibitem[Jin et~al.(2023)Jin, Yang, Chen, and Lu]{jin2023genegpt}
Qiao Jin, Yifan Yang, Qingyu Chen, and Zhiyong Lu.
\newblock Genegpt: Augmenting large language models with domain tools for improved access to biomedical information.
\newblock \emph{ArXiv}, 2023.

\bibitem[Kahneman \& Tversky(2000)Kahneman and Tversky]{Kahneman2000ChoicesVA}
D.~Kahneman and A.~Tversky.
\newblock Choices, values, and frames.
\newblock 2000.

\bibitem[Kahneman \& Tversky(2013)Kahneman and Tversky]{kahneman2013prospect}
Daniel Kahneman and Amos Tversky.
\newblock Prospect theory: An analysis of decision under risk.
\newblock In \emph{Handbook of the fundamentals of financial decision making: Part I}, pp.\  99--127. World Scientific, 2013.

\bibitem[Kocsis \& Szepesv{\'a}ri(2006)Kocsis and Szepesv{\'a}ri]{kocsis2006bandit}
Levente Kocsis and Csaba Szepesv{\'a}ri.
\newblock Bandit based monte-carlo planning.
\newblock In \emph{European conference on machine learning}, pp.\  282--293. Springer, 2006.

\bibitem[Lv et~al.(2024)Lv, Xia, and Huang]{lv2024codeact}
Weijie Lv, Xuan Xia, and Sheng-Jun Huang.
\newblock Codeact: Code adaptive compute-efficient tuning framework for code llms.
\newblock \emph{arXiv preprint arXiv:2408.02193}, 2024.

\bibitem[Maes(1994)]{Maes1994AgentsTR}
P.~Maes.
\newblock Agents that reduce work and information overload.
\newblock \emph{Commun. ACM}, 37:\penalty0 30--40, 1994.

\bibitem[Nakajima(2023)]{nakajima2023babyagi}
Yohei Nakajima.
\newblock Babyagi.
\newblock \emph{Python. https://github. com/yoheinakajima/babyagi}, 2023.

\bibitem[Nakano et~al.(2021)Nakano, Hilton, Balaji, Wu, Ouyang, Kim, Hesse, Jain, Kosaraju, Saunders, et~al.]{nakano2021webgpt}
Reiichiro Nakano, Jacob Hilton, Suchir Balaji, Jeff Wu, Long Ouyang, Christina Kim, Christopher Hesse, Shantanu Jain, Vineet Kosaraju, William Saunders, et~al.
\newblock Webgpt: Browser-assisted question-answering with human feedback.
\newblock \emph{ArXiv preprint}, abs/2112.09332, 2021.

\bibitem[OpenAI(2022)]{openaichatgptblog}
OpenAI.
\newblock Open{AI}: Introducing {ChatGPT}, 2022.
\newblock URL \url{https://openai.com/blog/chatgpt}.

\bibitem[OpenAI(2023)]{openai2023gpt4}
OpenAI.
\newblock Gpt-4 technical report, 2023.

\bibitem[Plott(1973)]{Plott1973PATHIR}
C.~Plott.
\newblock Path independence, rationality, and social choice.
\newblock \emph{Econometrica}, 41:\penalty0 1075--1091, 1973.

\bibitem[Qian et~al.(2024)Qian, Liu, Liu, Chen, Dang, Li, Yang, Chen, Su, Cong, et~al.]{qian2024chatdev}
Chen Qian, Wei Liu, Hongzhang Liu, Nuo Chen, Yufan Dang, Jiahao Li, Cheng Yang, Weize Chen, Yusheng Su, Xin Cong, et~al.
\newblock Chatdev: Communicative agents for software development.
\newblock In \emph{Proceedings of the 62nd Annual Meeting of the Association for Computational Linguistics (Volume 1: Long Papers)}, pp.\  15174--15186, 2024.

\bibitem[Qian et~al.(2023)Qian, Han, Fung, Qin, Liu, and Ji]{qian2023creator}
Cheng Qian, Chi Han, Yi~R Fung, Yujia Qin, Zhiyuan Liu, and Heng Ji.
\newblock Creator: Disentangling abstract and concrete reasonings of large language models through tool creation.
\newblock \emph{arXiv preprint arXiv:2305.14318}, 2023.

\bibitem[Qin et~al.(2023{\natexlab{a}})Qin, Cai, Jin, Yan, Liang, Zhu, Lin, Han, Ding, Wang, et~al.]{qin2023webcpm}
Yujia Qin, Zihan Cai, Dian Jin, Lan Yan, Shihao Liang, Kunlun Zhu, Yankai Lin, Xu~Han, Ning Ding, Huadong Wang, et~al.
\newblock Webcpm: Interactive web search for chinese long-form question answering.
\newblock \emph{arXiv preprint arXiv:2305.06849}, 2023{\natexlab{a}}.

\bibitem[Qin et~al.(2023{\natexlab{b}})Qin, Hu, Lin, Chen, Ding, Cui, Zeng, Huang, Xiao, Han, et~al.]{qin2023tool}
Yujia Qin, Shengding Hu, Yankai Lin, Weize Chen, Ning Ding, Ganqu Cui, Zheni Zeng, Yufei Huang, Chaojun Xiao, Chi Han, et~al.
\newblock Tool learning with foundation models.
\newblock \emph{arXiv preprint arXiv:2304.08354}, 2023{\natexlab{b}}.

\bibitem[Qin et~al.(2023{\natexlab{c}})Qin, Liang, Ye, Zhu, Yan, Lu, Lin, Cong, Tang, Qian, et~al.]{qin2023toolllm}
Yujia Qin, Shihao Liang, Yining Ye, Kunlun Zhu, Lan Yan, Yaxi Lu, Yankai Lin, Xin Cong, Xiangru Tang, Bill Qian, et~al.
\newblock Toolllm: Facilitating large language models to master 16000+ real-world apis.
\newblock \emph{arXiv preprint arXiv:2307.16789}, 2023{\natexlab{c}}.

\bibitem[Qin et~al.(2023{\natexlab{d}})Qin, Jagerman, Hui, Zhuang, Wu, Shen, Liu, Liu, Metzler, Wang, et~al.]{qin2023large}
Zhen Qin, Rolf Jagerman, Kai Hui, Honglei Zhuang, Junru Wu, Jiaming Shen, Tianqi Liu, Jialu Liu, Donald Metzler, Xuanhui Wang, et~al.
\newblock Large language models are effective text rankers with pairwise ranking prompting.
\newblock \emph{arXiv preprint arXiv:2306.17563}, 2023{\natexlab{d}}.

\bibitem[Richards(2023)]{richards2023auto}
Toran~Bruce Richards.
\newblock Auto-gpt: An autonomous gpt-4 experiment, 2023.

\bibitem[Schick et~al.(2023)Schick, Dwivedi-Yu, Dess{\`\i}, Raileanu, Lomeli, Zettlemoyer, Cancedda, and Scialom]{schick2023toolformer}
Timo Schick, Jane Dwivedi-Yu, Roberto Dess{\`\i}, Roberta Raileanu, Maria Lomeli, Luke Zettlemoyer, Nicola Cancedda, and Thomas Scialom.
\newblock Toolformer: Language models can teach themselves to use tools.
\newblock \emph{ArXiv preprint}, abs/2302.04761, 2023.

\bibitem[Searle(1969)]{Searle1969SpeechAA}
J.~Searle.
\newblock Speech acts: An essay in the philosophy of language.
\newblock 1969.

\bibitem[Sel et~al.(2023)Sel, Al-Tawaha, Khattar, Wang, Jia, and Jin]{sel2023algorithm}
Bilgehan Sel, Ahmad Al-Tawaha, Vanshaj Khattar, Lu~Wang, Ruoxi Jia, and Ming Jin.
\newblock Algorithm of thoughts: Enhancing exploration of ideas in large language models.
\newblock \emph{arXiv preprint arXiv:2308.10379}, 2023.

\bibitem[Shen et~al.(2023)Shen, Song, Tan, Li, Lu, and Zhuang]{shen2023hugginggpt}
Yongliang Shen, Kaitao Song, Xu~Tan, Dongsheng Li, Weiming Lu, and Yueting Zhuang.
\newblock Hugginggpt: Solving ai tasks with chatgpt and its friends in huggingface, 2023.

\bibitem[Shinn et~al.(2023)Shinn, Cassano, Labash, Gopinath, Narasimhan, and Yao]{shinn2023reflexion}
Noah Shinn, Federico Cassano, Beck Labash, Ashwin Gopinath, Karthik Narasimhan, and Shunyu Yao.
\newblock Reflexion: Language agents with verbal reinforcement learning, 2023.

\bibitem[Song et~al.(2023)Song, Xiong, Zhu, Li, Wang, Tian, and Li]{song2023restgpt}
Yifan Song, Weimin Xiong, Dawei Zhu, Cheng Li, Ke~Wang, Ye~Tian, and Sujian Li.
\newblock Restgpt: Connecting large language models with real-world applications via restful apis.
\newblock \emph{arXiv preprint arXiv:2306.06624}, 2023.

\bibitem[Sra et~al.(2011)Sra, Nowozin, and Wright]{sra2011optimization}
Suvrit Sra, Sebastian Nowozin, and Stephen~J Wright.
\newblock \emph{Optimization for machine learning}.
\newblock Mit Press, 2011.

\bibitem[Touvron et~al.(2023{\natexlab{a}})Touvron, Lavril, Izacard, Martinet, Lachaux, Lacroix, Rozi{\`e}re, Goyal, Hambro, Azhar, Rodriguez, Joulin, Grave, and Lample]{touvron2023llama}
Hugo Touvron, Thibaut Lavril, Gautier Izacard, Xavier Martinet, Marie-Anne Lachaux, Timoth{\'e}e Lacroix, Baptiste Rozi{\`e}re, Naman Goyal, Eric Hambro, Faisal Azhar, Aurelien Rodriguez, Armand Joulin, Edouard Grave, and Guillaume Lample.
\newblock Llama: Open and efficient foundation language models.
\newblock \emph{arXiv preprint arXiv:2302.13971}, 2023{\natexlab{a}}.

\bibitem[Touvron et~al.(2023{\natexlab{b}})Touvron, Martin, Stone, Albert, Almahairi, Babaei, Bashlykov, Batra, Bhargava, Bhosale, et~al.]{touvron2023llama2}
Hugo Touvron, Louis Martin, Kevin Stone, Peter Albert, Amjad Almahairi, Yasmine Babaei, Nikolay Bashlykov, Soumya Batra, Prajjwal Bhargava, Shruti Bhosale, et~al.
\newblock Llama 2: Open foundation and fine-tuned chat models.
\newblock \emph{arXiv preprint arXiv:2307.09288}, 2023{\natexlab{b}}.

\bibitem[Vemprala et~al.(2023)Vemprala, Bonatti, Bucker, and Kapoor]{vemprala2023chatgpt}
Sai Vemprala, Rogerio Bonatti, Arthur Bucker, and Ashish Kapoor.
\newblock Chatgpt for robotics: Design principles and model abilities.
\newblock Technical Report MSR-TR-2023-8, Microsoft, February 2023.

\bibitem[Wang et~al.(2023{\natexlab{a}})Wang, Li, Chen, Zhu, Lin, Cao, Liu, Liu, and Sui]{wang2023large}
Peiyi Wang, Lei Li, Liang Chen, Dawei Zhu, Binghuai Lin, Yunbo Cao, Qi~Liu, Tianyu Liu, and Zhifang Sui.
\newblock Large language models are not fair evaluators.
\newblock \emph{arXiv preprint arXiv:2305.17926}, 2023{\natexlab{a}}.

\bibitem[Wang et~al.(2023{\natexlab{b}})Wang, Cai, Chen, Liu, Ma, and Liang]{wang2023describe}
Zihao Wang, Shaofei Cai, Guanzhou Chen, Anji Liu, Xiaojian Ma, and Yitao Liang.
\newblock Describe, explain, plan and select: Interactive planning with large language models enables open-world multi-task agents.
\newblock \emph{arXiv preprint arXiv:2302.01560}, 2023{\natexlab{b}}.

\bibitem[Wei et~al.(2023)Wei, Wang, Schuurmans, Bosma, Ichter, Xia, Chi, Le, and Zhou]{wei2023chainofthought}
Jason Wei, Xuezhi Wang, Dale Schuurmans, Maarten Bosma, Brian Ichter, Fei Xia, Ed~Chi, Quoc Le, and Denny Zhou.
\newblock Chain-of-thought prompting elicits reasoning in large language models, 2023.

\bibitem[Wooldridge \& Jennings(1995)Wooldridge and Jennings]{Wooldridge1995IntelligentAT}
M.~Wooldridge and N.~Jennings.
\newblock Intelligent agents: theory and practice.
\newblock \emph{The Knowledge Engineering Review}, 10:\penalty0 115 -- 152, 1995.

\bibitem[Wu et~al.(2023{\natexlab{a}})Wu, Yin, Qi, Wang, Tang, and Duan]{wu2023visual}
Chenfei Wu, Shengming Yin, Weizhen Qi, Xiaodong Wang, Zecheng Tang, and Nan Duan.
\newblock Visual chatgpt: Talking, drawing and editing with visual foundation models.
\newblock \emph{ArXiv preprint}, abs/2303.04671, 2023{\natexlab{a}}.

\bibitem[Wu et~al.(2023{\natexlab{b}})Wu, Bansal, Zhang, Wu, Zhang, Zhu, Li, Jiang, Zhang, and Wang]{wu2023autogen}
Qingyun Wu, Gagan Bansal, Jieyu Zhang, Yiran Wu, Shaokun Zhang, Erkang Zhu, Beibin Li, Li~Jiang, Xiaoyun Zhang, and Chi Wang.
\newblock Autogen: Enabling next-gen llm applications via multi-agent conversation framework.
\newblock \emph{arXiv preprint arXiv:2308.08155}, 2023{\natexlab{b}}.

\bibitem[Yang et~al.(2023)Yang, Chen, Li, Ding, and Wu]{yang2023chatgpt}
Linyao Yang, Hongyang Chen, Zhao Li, Xiao Ding, and Xindong Wu.
\newblock Chatgpt is not enough: Enhancing large language models with knowledge graphs for fact-aware language modeling.
\newblock \emph{arXiv preprint arXiv:2306.11489}, 2023.

\bibitem[Yao et~al.(2022{\natexlab{a}})Yao, Chen, Yang, and Narasimhan]{yao2022webshop}
Shunyu Yao, Howard Chen, John Yang, and Karthik Narasimhan.
\newblock Webshop: Towards scalable real-world web interaction with grounded language agents.
\newblock \emph{Advances in Neural Information Processing Systems}, 35:\penalty0 20744--20757, 2022{\natexlab{a}}.

\bibitem[Yao et~al.(2022{\natexlab{b}})Yao, Zhao, Yu, Du, Shafran, Narasimhan, and Cao]{yao2022react}
Shunyu Yao, Jeffrey Zhao, Dian Yu, Nan Du, Izhak Shafran, Karthik Narasimhan, and Yuan Cao.
\newblock React: Synergizing reasoning and acting in language models.
\newblock \emph{ArXiv preprint}, abs/2210.03629, 2022{\natexlab{b}}.

\bibitem[Yao et~al.(2023)Yao, Yu, Zhao, Shafran, Griffiths, Cao, and Narasimhan]{yao2023tree}
Shunyu Yao, Dian Yu, Jeffrey Zhao, Izhak Shafran, Thomas~L Griffiths, Yuan Cao, and Karthik Narasimhan.
\newblock Tree of thoughts: Deliberate problem solving with large language models.
\newblock \emph{arXiv preprint arXiv:2305.10601}, 2023.

\end{thebibliography}
\bibliographystyle{iclr2025_conference}

\appendix
\section{Prompt Design}
\label{sec:prompt}

\subsection{Utility Learning Prompt}
Our utility learning prompt is designed as follows:
\begin{lstlisting}[basicstyle=\ttfamily, breaklines=true]
You are value-GPT, an expert in defining which trail is better and closer to solving the task. Here is the task description:
*******************************
{{BEGIN_DESCRIPTION}}
your_task: {task_description}
your_query: {input_description}
{{END_DESCRIPTION}}
*******************************
Here are two candidates A and B. They both try to handle the task with some function calls. Their trails are as follows.
*******************************
{{CANDIDATE_A_START}}
{candidate_A}
{{CANDIDATE_A_END}}
*******************************
{{CANDIDATE_B_START}}
{candidate_B}
{{CANDIDATE_B_END}}
*******************************
\end{lstlisting}

Then, ChatGPT should call the following function\footnote{https://openai.com/blog/function-calling-and-other-api-updates} to give the result.
\begin{lstlisting}[basicstyle=\ttfamily, breaklines=true]
{
    "name": "choose_preference",
    "description": "Choose the preferred answer for the query within all given answers.",
    "parameters": {
        "type": "object",
        "properties": {
            "preference": {
                "type": "number",
                "description": "The index of the preferred answer in all given answers."
            },
        },
    },
}
\end{lstlisting}

\subsection{Decision-Making Prompt}
Our decision-making prompt is designed as follows:
\begin{lstlisting}[basicstyle=\ttfamily, breaklines=true]
You are the Decision-Making GPT and can perform any task using the tree search method. 
The search method is as follows:
1. First, I will provide you with the task description and input details.
2. For each task, you need to call various functions through multiple steps. 
3. At each step, you need to give your thought to analyze the status now and what to do next, with a function call to actually execute your step.
After the call, you will get the call result, and you are now in a new state.
4. Each (thought-function) pair mentioned above is considered a tree node, and each trail is a tree path from the root to a terminal node. Therefore, the Monte Carlo search tree contains multiple trails.
5. Although you may not see previous trails, in each trail, I will first place you in an intermediate state determined by the "value" of the node, and then you make different choices from there.

Remember:
1. Always make a function call at each step.
2. If you believe you have gathered enough information, call the function "Finish: give_answer" to provide your answer for the task.
3. If you feel unable to handle the task from this step, call the function "Finish: give_up_and_restart".

Let's begin!
Task description: {task_description}
\end{lstlisting}

\subsection{Task Description}
\label{sec:task_description}
The task description of Game of 24 is as follows:
\begin{lstlisting}[basicstyle=\ttfamily, breaklines=true]
Use numbers and basic arithmetic operations (+ - * /) to obtain exactly one number 24. In each step, you are only allowed to choose two of the left numbers to obtain a new number. For example, 7 * 9 - 3 * 1 3 = 2 4.
Remember:
1. All of the numbers must be used and must be used ONCE. So Only when the left number is exactly 24, you will win. So you don't succeed when the left number = [2 4, 5]. You succeed when the left number = [2 4]. 
2. All the try takes exactly 3 steps, and the task ends when the count of left numbers is 1, and check whether the only left number equals 24.
3. When there are two numbers left, ALWAYS pre-compute and list all the operations' combined results( + - * / ), and find if there is a way to combine them to 24 before you make the function call. 
3.1. If There is a way, use function "play_24" to combine it.
3.2. If not, use function "give_up" to restart.
4. The status changes ONLY when you call function "play_24". If you only give thoughts, nothing happens.
5. "play_24" inputs only one step, if you want to combine many numbers, split it into multiple calls. 
\end{lstlisting}

The task description of Webshop is listed as follows:
\begin{lstlisting}[basicstyle=\ttfamily, breaklines=true]
You should use functions to help handle the web shopping task on a webshop site.
We have 2 Pages: Product Selection Page & Product Details Page.
You have access to the following functions:
1. search: at any time, you can search a product by keywords. Then you will go to the Product Selection Page which shows a list of related products.
2. select: after searching keywords, you can select a product on the Product Selection Page. Then you will go to the Product Details Page, which shows the details of the product you select.
3. buy: On the Product Details Page, you can buy a product. Then the shopping task is completed.
\end{lstlisting}

The task description of ToolBench is listed as follows:
\begin{lstlisting}[basicstyle=\ttfamily, breaklines=true]
You should use functions to help handle real-time user queries. Remember:
1. ALWAYS call "Finish" function at the end of the task. And the final answer should contain enough information to show to the user, If you can't handle the task, or you find that function calls always fail(the function is not valid now), use function Finish->give_up_and_restart.
2. Do not use origin tool names, use only subfunctions' names.
You have access to the following tools:
\end{lstlisting}

\subsection{Simple Prompt of Utility Learning}
\label{sec:simple_prompt}
\begin{lstlisting}[basicstyle=\ttfamily, breaklines=true]
    Giving task description and candidate answers, I want you to choose one preferred answer which is more close to success.
    
    *******************************
    {{BEGIN_DESCRIPTION}}
    your_task: {task_description}
    your_query: {input_description}
    {{END_DESCRIPTION}}
    *******************************
    
    *******************************
    {{CANDIDATE_0_START}}
    {candidate_A}
    {{CANDIDATE_0_END}}
    *******************************
    {{CANDIDATE_1_START}}
    {candidate_B}
    {{CANDIDATE_1_END}}
    *******************************
\end{lstlisting}




\section{Hyperparameter Analysis}
\label{sec:hyperparameter}

\subsection{Selection of Elo coeficient}
The selection of $r$ is based on the classic Elo rating algorithm. 
In the classic Elo rating system~\citep{elo1978rating}, the expected superiority is defined as:
\begin{equation}
    E_{x>y} = \frac{1}{1 + 10^{-\frac{v_{x} - v_{y}}{400}}}
\end{equation}
In this paper, we employ the base to compute the expected superiority for computational implementation convenience:
\begin{equation}
    E_{x>y} = \frac{1}{1 + e^{-\frac{v_{x} - v_{y}}{r}}}
\end{equation}
To achieve equivalence between them, we can set as approximately $r = \frac{400}{\ln 10} \approx 173.72$ to change the base. 
In this way, our calculation of the expected superiority equals the classic Elo rating algorithm.

\subsection{Impact of initial Elo Score}
As Equation~\ref{eqn:Elo_expect} shows, within the Elo rating algorithm, the decision-making process is influenced by the relative difference between Elo scores of decision steps, rather than their absolute values.
The essence of the Elo rating system lies in its dynamic nature, where scores are adjusted based on pairwise comparisons over time. 
This means that regardless of the initial score assigned to each decision step, the subsequent adjustments made through pairwise comparisons are what determine the final, accurate assessment of each decision's utility. 
Therefore, the initial score primarily serves as a starting point, and its specific value is not critical to the overall decision-making process.
Consequently, the initial Elo score assigned to decision steps does not fundamentally impact the outcome of the comparisons. 

\begin{table}[!h]
    \centering
    \caption{Success rate of different Elo score on Game of 24.}
    \vskip 0.1in
    \begin{tabular}{lc}
    \toprule
    Value of Elo & Success Rate \\
    \midrule
    Elo$=0$ & 62.0 \\
    Elo$=100$ & 60.0 \\
    Random Elo & 63.0 \\
    \bottomrule
    \end{tabular}
    \label{tab:perf_of_diff_elo}
\end{table}

To validate the reliability of our Elo rating system, we further conducted two experiments. 
First, we manualy initialize Elo score of all decision steps as $100$ instead of $0$ in our original settings. 
Second, all Elo scores are initialized randomly so each decision step has different Elo scores in the begining. 
We re-run our method in the Game of 24 scenario and assess the final performance. 
The results are listed in \Cref{tab:perf_of_diff_elo} and it can find that different initialized Elo scores result in the similar final performance of our method. 
The results show that despite the different initial values, the final performance remains consistent. 
This consistency confirms that through multiple comparisons, Elo scores converge to their true value regardless of their initialization, thus validating the robustness of our approach.
Note that we implement our method and baseline based on GPT-4o-mini to reduce the API cost so the performance does not equal those reported in the main paper.

\subsection{Impact of the Converge of Elo Rating}

As the Elo score will converge after multiple comparisons, the number of comparisons may impact the final performance.
In our experiments, we set the number of comparison in the utility learning as $2$ per candidates, i.e., each candidate decision sequence should be compared twice.
To validate the impact of the number of comparison, we double it as $4$ and re-run the experiments on Game of 24.
The results are listed in \Cref{tab:perf_of_diff_comparison} and we find that doubling the number of the comparison cannot bring improvement, revealing that Elo scores have converged to their true values.
Note that we implement our method based on GPT-4o-mini to reduce the API cost so the performance does not equal those reported in the original paper. 
\begin{table}[!h]
    \centering
    \caption{Success rate of different the number of comparison on Game of 24.}
    \vskip 0.1in
    \begin{tabular}{lc}
    \toprule
    Settings & Success Rate \\
    \midrule
    \#.CMP $=2$ & 62.0 \\
    \#.CMP $=4$ & 62.0 \\
    \bottomrule
    \end{tabular}
    \label{tab:perf_of_diff_comparison}
\end{table}

\subsection{Impact of Initial Decision Solution}

As the utility learning are conducted after exploring multiple decision sequences for comparisons, the first explored decision sequence (i.e., initial deicision sequence) may impact the utility learning effectiveness and further influence the exploration process.
To validate the robustness of our method on the initial decision sequence, we conducted additional experiments using the Game of 24 scenario.
Given that the Game of 24 requires three-step decisions, we manually initialized three different decision sequences for each instance, where the first, second, and third decisions were deliberately set as the incorrect decisions. 
This allowed us to test our method's resilience across different initial conditions. 
The experiment results are listed in \Cref{tab:perf_of_diff_initial_solution},
\begin{table}[!t]
    \centering
    \caption{Success rate of different initial decision solution on Game of 24.}
    \vskip 0.1in
    \begin{tabular}{lc}
    \toprule
    Settings & Success Rate \\
    \midrule
    Init. w/ Incorrect First Step & 59.0 \\
    Init. w/ Incorrect Second Step & 61.0 \\
    Init. w/ Incorrect Third Step & 60.0 \\
    \bottomrule
    \end{tabular}
    \label{tab:perf_of_diff_initial_solution}
\end{table}
Despite the different initial decision sequences, our method consistently achieved similar success rates, demonstrating its robustness. 
Note that we implement our method and baseline based on GPT-4o-mini to reduce the API cost so the performance does not equal those reported in the main paper.

The Elo rating system is specifically designed to be resilient~\citep{elo1967proposed}, even when initial decision policies are suboptimal. 
This is achieved by continuously comparing and updating the Elo scores accordingly, allowing the Elo scores to progressively converge to their true values. 
Adapting the Elo rating system in our method, it can provide reliable measure the utility of each decision step. 
Even if the initial decision sequence is suboptimal, through iteratively experience exploration and utility learning, each decision step would be measured accurately and our method would avoid the suboptimal and even poor decisions. 
As a result, the effectiveness of our method does not heavily depend on the quality of the initial decision sequences. 
Instead, it is capable of learning and improving from its experiences, adapting to better strategies as it gains more insights through repeated interactions.
These results further confirm that our method can effectively adapt and converge to optimal decision paths, even when starting from suboptimal sequences.

\subsection{Efficiency Analysis on Game of 24}
\label{sec:efficiency_on_24}

To further validate the generalized efficiency of our method, we further conducted efficiency experiments in the Game of 24 scenario. 
Similar to the settings in \Cref{sec:efficiency}, we manualy set different API call budgets and assess the final performance of our method against the best baseline, DFSDT. 
The results are listed in \Cref{tab:efficiency_on_24}. 
Obviously, we can find that the similar efficiency results show in the Game of 24.
Our method still achieve Highest Performance for Same Cost and Lowest Cost for Same Performance. 
Such results further validates the efficiency superiority of our method against baselines.
Note that we implement our method and baseline based on GPT-4o-mini to reduce the API cost so the performance does not equal those reported in the main paper.

\begin{table}[!h]
    \centering
    \caption{Efficiency analysis of different methods on Game of 24.}
    \vskip 0.1in
    \begin{tabular}{lcccc}
    \toprule
    \multirow{2}*{Model} & \multicolumn{4}{c}{API call budget} \\
     & 50 & 100 & 150 & 200 \\
    \midrule
    DFSDT & 16.0 & 32.0 & 35.0 & 42.0 \\
    \model & 20.0 & 38.0 & 52.0 & 55.0 \\
    \bottomrule
    \end{tabular}
    \label{tab:efficiency_on_24}
\end{table}


\subsection{Generalizability to Various Large Language Models}
To verify the generalizability of \model to various LLMs including open-source LLMs, we have conducted additional experiments using two alternative representative LLMs: the proprietary \texttt{claude-3-5-haiku-20241022} and the open-source \texttt{llama-3.1-8b}. We re-implemented \model with these models and evaluated its performance on the Game of 24 task. The experimental results are presented in the \Cref{fig:various_llms}. The results demonstrate that \model continues to yield significant performance improvements when integrated with different LLMs, underscoring the robustness and generalizability of our proposed approach.

\begin{table}[h]
    \centering
    \caption{Performance based on various LLMs on Game of 24.}
    \vskip 0.1in
    \begin{tabular}{lcc}
        \toprule
        LLM & CoT@1 & RaDAgent \\
        \midrule
        llama-3.1-8b & 3.0 & 15.0 \\
        claude-3-5-haiku & 2.0 & 71.0 \\
        gpt-3.5-turbo & 6.0 & 43.0 \\
        \bottomrule
    \end{tabular}
    \label{fig:various_llms}
\end{table}

\subsection{Error Analysis}


In this section, we present comprehensive analysis to show the failure in decision-making in ToolBench.
We commence our analysis by categorizing the common reasons for failure encountered by each model in ToolBench. 
These reasons encompass:
(1) \textbf{Tool Inaccessibility}: Occurrences where a subset of the designated tools is inaccessible, e.g., HTTP 404 or 500 error.
(2) \textbf{Parameter Error}: Occurrences when call tools, including parameter format mismatching and missing mandatory parameter fields.
(3) \textbf{Tool Hallucination}: Instances where the model employs tools not provided, i.e., invoking a non-existent tool.
(4) \textbf{Decision Failure}: Instances where the model fails to accomplish although none of the aforementioned problems occur.
We present the incidence ratio of the aforementioned categories.
Specifically, the incidence ratios are calculated based on the entire exploration process. 
If the error occurs in the explored decision tree (even not in the final decision sequence), it is still counted.
As Tool Hallucination and Parameter Error can be fixed during decision making, we further report the fix ratio that agents successfully accomplish the instructions although errors occurred.

\begin{table*}[!h]
    \small
    \centering
    \caption{Incidence ratio and Fix ratio of common failure reasons in ToolBench dataset.}
    \vskip 0.1in
    \begin{tabular}{lcccccc}
    \toprule
    \multicolumn{1}{l}{\multirow{2}{*}{Method}} & \multicolumn{2}{c}{Tool Hallucination} & \multicolumn{2}{c}{Parameter Error} & \multirow{2}{*}{\shortstack[c]{Tool Inaccessibility}} & \multirow{2}{*}{\shortstack[c]{Decision Failure}} \\
    \cmidrule{2-5}
    \multicolumn{1}{c}{}                        & Incidence              & Fix          & Incidence          & Fix         & &  \\
    \midrule
    CoT@3                                       & 14.2               & 25.4                   & 41.2     & 14.8      & 2.0                                 & 52.5          \\
    BFS                                         & 18.8               & 25.5                   & 50.8   & 31.1                & 2.6                                 & 48.6       \\
    DFSDT                                       & 31.5              & 38.9                  & 62.5                & 41.0             & 3.0                                 & 26.4     \\
    \model                                      & 42.1               & \textbf{53.3}       & 62.3   & \textbf{54.0}           & 3.0     & \textbf{14.8} \\
    \bottomrule
    \end{tabular}
    \label{tab:common_failure_reason}
\end{table*}

From Table~\ref{tab:common_failure_reason}, several noteworthy observations arise:
(1) \model boasts the lowest incidence ratio of decision failure, highlighting its adeptness in decision making.
(2) As \model conducts a diverse and extensive exploration, it will experience more parameter errors and tool hallucinations, causing a higher incidence ratio. 
This diverse exploration is integral as it allows \model to thoroughly evaluate a wide array of possible decision pathways, even those that are less conventional or more prone to errors.
Despite exploring riskier or more error-prone decisions, \model effectively utilizes the Elo-based utility learning mechanism to learn from diverse explorations and subsequently pinpoint the most efficient and error-free pathway. 
The high fix ratio underlines \model's ability to rectify potential errors encountered during the exploration phase, ultimately leading to a reliable and effective decision-making process.
%
%
(3) All methods own similar incident ratio of tool inaccessibility which shows that there still exist some inoperative APIs in ToolBench, influencing the decision-making process.
(4) We examine cases that all methods fail and find certain cases remain unsolvable due to the ambiguity of user-provided values (e.g., user ID, email address) or restrictions imposed by limited tool chain lengths, which underscores the necessity for advanced decision-making proficiencies.

\section{Broader Impact}
\label{sec:impact}


This paper presents work whose goal is to advance the field of Autonomous Decision-Making for Large Language Models. 
There are many potential broader impacts of our work and we discuss some aspects in the following:
Firstly, our work explores the internal rationality for LLM-based agents, and the proposed \elo could leverage Elo Rating system to construct quantitative utilities for each decisions. 
This advancement could have profound implications for industries where decision-making is critical, such as finance, healthcare, and law. 
However, this autonomy raises ethical considerations regarding the extent to which LLMs should be allowed to make decisions without human oversight, especially in high-stakes scenarios.
Secondly, as \model demonstrates improved efficiency and cost-effectiveness in decision-making tasks, there could be an increased reliance on LLMs in various sectors. 
This dependence may lead to a reduction in human labor for certain tasks and could influence the job market, necessitating a re-evaluation of workforce skills and training.
Thirdly, while \model learns from its posterior experiences, the quality and diversity of these experiences are crucial. 
There is a potential risk of inheriting biases present in the training data or developing new biases based on limited or skewed experiences. 
This aspect necessitates continuous monitoring and updating of the model to ensure fairness and impartiality in decision-making.

\section{Limitations}
Our approach still has several limitations:
(1) Our method involves exploring new decision traces from intermediate decision steps, necessitating the recovery of the state at each step. 
In practice, certain decisions cannot be reversed once executed. 
In these instances, our method requires a sophisticated rollback mechanism to function correctly.
(2) Our utility learning method relies on the comparative judgment capabilities of large language models (LLMs) to achieve Elo ratings. 
While GPT-3.5 and GPT-4 can implement our method, it is uncertain if other LLMs, especially for open-source LLMs, can achieve similar performance. 
We will explore these limitations in the future.

\end{document}